\documentclass[journal]{IEEEtran}

\usepackage{cite}
\usepackage{amsmath,amssymb,amsfonts}
\usepackage{graphicx}
\usepackage{textcomp}
\usepackage{xcolor}
\usepackage{booktabs}
\usepackage{multirow}
\usepackage{url}
\usepackage{enumitem}
\usepackage{hyperref}
\usepackage{orcidlink}

\hypersetup{
  colorlinks=true,
  linkcolor=blue,
  citecolor=blue,
  urlcolor=blue
}

\setlist[itemize]{leftmargin=*,topsep=2pt,itemsep=1pt,parsep=1pt}

\title{UoU: A Universal Fingerprint Foundation Model Based on Large-Scale Unsupervised Learning}
\author{Xiongjun~Guan$^{\orcidlink{0000-0001-8887-3735}}$,
Jianjiang~Feng$^{\orcidlink{0000-0003-4971-6707}}$,~\IEEEmembership{Member,~IEEE},
and Jie~Zhou$^{\orcidlink{0000-0001-7701-234X}}$,~\IEEEmembership{Senior Member,~IEEE}%
\thanks{Xiongjun Guan, Jianjiang Feng, and Jie Zhou are with the Department of Automation, Tsinghua University, Beijing 100084, China (e-mail: \url{gxj21@mails.tsinghua.edu.cn}; \url{jfeng@tsinghua.edu.cn}; \url{jzhou@tsinghua.edu.cn}).}}

\begin{document}
\maketitle

\begin{abstract}
Fingerprint recognition is still dominated by task-specific pipelines, where enhancement, structural parsing, alignment, and matching are optimized in isolation. Although effective in narrow settings, this design limits representation reuse across sensors, qualities, and downstream applications. We therefore present UoU, short for ``a \textbf{U}niversal fingerprint foundation model based \textbf{o}n large-scale \textbf{U}nsupervised learning,'' which reframes fingerprint feature extraction as a domain-specific foundation-model problem. UoU is organized around a multi-level representation hierarchy spanning image restoration, structural fields, semantic tokens, point-level biometric entities, and compact global descriptors. Its training recipe combines a supervised cold start on precise annotations, large-scale weakly supervised refinement, and large-scale unsupervised consolidation, with the latter two stages iterated during large-scale training so that weak supervision broadens semantic coverage while unsupervised learning stabilizes correspondences, invariances, and representation geometry. Rather than treating fingerprint imagery as generic texture, UoU exploits domain-specific symmetries and intermediate structure, including orientation flow, periodic ridge patterns, sparse biometric entities, and spatial equivariance. The framework is intentionally architecture-agnostic: while the present study includes an initial transformer-based structured-prediction instantiation, the broader design supports multi-task learning, scalable model configurations, and downstream specialization for matching, alignment, enhancement, registration, and related fingerprint applications. This paper presents the technical motivation, system design, and validation protocol of UoU, and part of the baseline implementation is publicly available at \url{https://github.com/XiongjunGuan/UoU}.
\end{abstract}


\section{Introduction}
Fingerprint analysis remains a core technology for identity recognition, forensics, access control, and trusted human-computer interaction. Yet many practical systems still decompose the problem into separate stages, such as enhancement, structural parsing, matching, and quality-aware decision making. While this decomposition is historically effective, it also creates brittle interfaces between modules and encourages features that are optimized for one subproblem but not readily transferable to others.

Recent progress in representation learning suggests a different route: rather than designing isolated task-specific algorithms, one can seek a shared backbone that learns reusable representations across heterogeneous supervision sources and downstream objectives. This shift has transformed general vision through transformers, self-supervised learning, and foundation-style pretraining~\cite{vaswani2017attention,he2016resnet,carion2020detr,dosovitskiy2021vit,he2022mae}. However, the fingerprint domain still lacks a mature counterpart that unifies pixel-level restoration, field-level geometry, token-level abstraction, point-level biometrics, and global identity-aware descriptors within a single scalable framework. Existing deep fingerprint systems have achieved strong progress in structural extraction, compact matching, pose alignment, dense registration, and latent comparison~\cite{tang2017fingernet,nguyen2018minutiaenet,engelsma2019deepprint,pan2025fixedlengthdense,guan2024jointpose,guan2025underpose,guan2024distortionfield,guan2024pdrnet,pan2024dmd}, but they remain largely optimized task by task rather than as instances of a transferable foundation-model paradigm.

This gap is becoming increasingly consequential because fingerprint deployment scenarios are expanding much faster than the representational assumptions of legacy pipelines. Under-screen sensors, partial impressions, latent traces, mobile capture, spoof resistance, and large-scale search all stress different aspects of fingerprint understanding, including denoising, correspondence, deformation tolerance, and identity discrimination. A model that is excellent at one stage of one pipeline may still fail to transfer to the neighboring stage or to a new capture regime. What the field now needs is not only stronger task solvers, but a reusable fingerprint foundation that can support them jointly.

From this perspective, the central question is not whether a single module can be improved in isolation, but whether fingerprint analysis can be organized around a reusable modeling paradigm that absorbs multiple forms of supervision and remains extensible as architectures evolve. Such a paradigm should preserve structural priors such as orientation, singular points, and deformation cues; support semantic abstraction through tokens or compact descriptors; accommodate multiple deployment scales; and enable downstream specialization through fine-tuning or lightweight adaptation. This is the level at which we position UoU: a universal fingerprint foundation-model paradigm organized around a representation hierarchy and a shared backbone reusable across tasks. Although the title emphasizes large-scale unsupervised learning, the intended meaning is not that every stage is annotation-free; rather, unsupervised or weakly supervised fingerprint data provide the principal scaling substrate, while precise supervision anchors semantic structure and iterative unsupervised consolidation refines it at scale.

To address this gap, we develop UoU together with an initial implementation. The present study already provides a meaningful first step in the form of a transformer-based structured-prediction branch for fingerprint-aware geometric understanding. This branch is not intended to define the boundary of UoU, but to serve as an initial instantiated component within a broader framework. Its role is to demonstrate a feasible entry point through which fingerprint-aware structural prediction can be converted into reusable backbone features for more general analysis tasks.

We position this work as a foundation-model study for universal fingerprint representation learning with an initial implemented branch. The broader UoU design is centered on large-scale unsupervised or weakly supervised fingerprint resources as the main source of breadth, but it does not treat those resources as sufficient by themselves. Instead, the shared backbone is first initialized by a small amount of precise, high-quality supervision, then expanded by large-scale weak supervision, and finally consolidated through large-scale unsupervised learning on broader fingerprint corpora. In practice, the latter two stages are not best viewed as a single forward pass through a fixed pipeline; they are better understood as an iterative loop in which weak supervision broadens semantic coverage and unsupervised learning stabilizes representations, correspondences, and invariances at scale. This direction is particularly well suited to fingerprint analysis because the domain naturally provides multiple intermediate signals, including ridge orientation, periodic texture, singular structures, local geometry, masks, local patches, deformation cues, and global identity-sensitive descriptors. Fingerprints are therefore not merely another visual domain; they constitute a structured biometric domain with unusually rich intermediate supervision and a strong case for reusable domain representations.

Under this framing, the manuscript serves two roles. It introduces an initial but technically coherent structured branch, and it formalizes the broader UoU design as a concrete architecture and training roadmap for universal fingerprint foundation modeling.

The intended message of the paper is therefore broader than a narrow single-task model. Rather than treating enhancement, structure extraction, alignment, and recognition as isolated engineering products, UoU argues that they should be interpreted as downstream views of one shared fingerprint foundation. The contribution of this work is to make that claim technically explicit through a representation hierarchy, a data-quality-aware large-scale training strategy, and an initial implemented branch that grounds the framework in a concrete learning system.

\noindent\textbf{Contributions.} The main contributions of this work are summarized as follows:
\begin{itemize}
\item We introduce UoU as a universal fingerprint foundation-model paradigm centered on reusable domain representations rather than isolated task endpoints.
\item We establish a representation hierarchy spanning image, field, token, point, and global levels, thereby linking fingerprint-specific structure to scalable downstream adaptation.
\item We formulate an architecture-agnostic large-scale training strategy that combines supervised cold start, weakly supervised refinement, and iterative unsupervised consolidation for universal fingerprint modeling.
\item We present a concrete initial instantiation that grounds the framework and supports subsequent empirical extension.
\end{itemize}

\section{Related Work}
\textbf{Fingerprint representation and matching.}
Classical fingerprint systems build around handcrafted ridge processing, singular point estimation, minutiae extraction, and matcher-specific templates. These pipelines remain influential because they exploit strong biometric priors, as extensively summarized in fingerprint recognition literature~\cite{maltoni2009handbook}. Public evaluations such as FVC have also shaped the field by emphasizing reproducible benchmarking under realistic acquisition variability~\cite{cappelli2006fvc2004}. Nevertheless, modular pipelines can struggle when the image is low quality, heavily deformed, or only partially observed.

\textbf{Minutiae, orientation, and structured feature extraction.}
Modern fingerprint learning increasingly replaces handcrafted stages with neural modules for enhancement, minutiae localization, and representation learning. Early CNN-based latent minutiae pipelines already showed that learned feature maps can outperform hand-engineered detectors under severe background noise~\cite{tang2017latentfcn}. Later systems such as FingerNet~\cite{tang2017fingernet} and MinutiaeNet~\cite{nguyen2018minutiaenet} more explicitly fused fingerprint priors with learned modules, demonstrating the value of combining domain structure and deep networks. FingerNet and DeepPrint~\cite{engelsma2019deepprint} are especially relevant here because they do not only support minutiae-aware processing, but also expose the importance of orientation, alignment, and explicit fingerprint structure in representation learning. However, these methods are still primarily structured around a particular task endpoint, whereas UoU is explicitly framed around a reusable representation hierarchy that should survive task changes.

\textbf{Fixed-length fingerprint representations.}
Another important line of work seeks compact descriptors for fast retrieval and recognition. DeepPrint~\cite{engelsma2019deepprint} is a representative example that learns a fixed-length fingerprint representation while still incorporating domain knowledge such as alignment and minutiae cues. More recently, fixed-length dense representations have been studied to preserve richer local structure while maintaining compactness and efficient comparison~\cite{pan2025fixedlengthdense}. These methods demonstrate that fixed-length learned representations are powerful, but they still target a narrower representational endpoint than UoU, which seeks to unify compact descriptors with structural fields, tokens, and point-level features under one encoder.

\textbf{Pose estimation and alignment.}
Pose variation is a critical factor in partial and mobile fingerprint scenarios. Recent representative works include joint identity verification and pose alignment for partial fingerprints~\cite{guan2024jointpose} and finger pose estimation for under-screen fingerprint sensors~\cite{guan2025underpose}. These works show that pose estimation is not merely a preprocessing step, but is deeply coupled with downstream recognition quality. UoU differs in emphasis by treating such geometry-sensitive reasoning as one branch that should be absorbed into a universal fingerprint foundation model, rather than as a standalone specialization.

\textbf{Dense deformation field estimation and registration.}
Beyond global pose, dense non-rigid distortion remains a major challenge in real fingerprint acquisition. Regression of dense distortion fields from a single fingerprint image~\cite{guan2024distortionfield} and phase-aggregated dual-branch dense registration~\cite{guan2024pdrnet} are representative efforts that move from coarse alignment to fine deformation modeling. Their relevance to UoU is twofold: they emphasize the need for geometry-sensitive representations, and they motivate future extensions from 2D pose reasoning to richer deformation-aware universal modeling. In contrast to registration-specific systems, UoU aims to treat deformation reasoning as one reusable capability within a larger fingerprint representation program.

\textbf{Enhancement, restoration, and generation.}
Image enhancement remains a foundational problem because many downstream fingerprint tasks degrade sharply when ridge clarity is poor. FingerNet~\cite{tang2017fingernet} is again representative because it integrates enhancement-related processing with downstream structured extraction. PrintsGAN~\cite{engelsma2022printsgan}, while framed as synthetic generation, is also highly relevant because it demonstrates how data generation and realism can support representation learning at scale. In the broader UoU perspective, enhancement and generation are not isolated tasks; they are natural branches of a shared fingerprint framework, and should therefore be connected through the same encoder rather than through disconnected pipelines.

\textbf{Transformers and query-based structured prediction.}
Query-based set prediction has shown that structured visual targets can be modeled as unordered sets with learned queries and bipartite assignment~\cite{carion2020detr}. This formulation is especially appealing here because many fingerprint structural annotations can be expressed with variable cardinality and weak ordering assumptions. The present implementation adopts this general perspective for fingerprint-aware structured prediction. At a lower level, the design also inherits from convolutional backbones~\cite{he2016resnet} and transformer attention~\cite{vaswani2017attention}.

\textbf{Foundation-style representation learning.}
Large-scale self-supervised and masked modeling methods have highlighted the value of pretraining shared backbones before downstream adaptation~\cite{dosovitskiy2021vit,he2022mae}. UoU adopts the same philosophy, but specializes it to fingerprint data, where domain priors can define richer token semantics than generic visual patches alone. Relative to generic pretraining, the central question here is how to organize fingerprint-specific supervision and invariance into a reusable latent space. This is precisely the gap that remains open between strong task-level fingerprint networks and a truly universal fingerprint foundation model.

Across these lines of work, a common pattern emerges: the field already contains strong modules for enhancement, structure extraction, alignment, compact matching, and deformation reasoning, but these modules are rarely organized as different views of one shared fingerprint intelligence system. This observation motivates the role of UoU in the literature. Rather than competing with each prior line on a single endpoint, UoU aims to provide a unifying foundation-model perspective in which these task families become coordinated consumers or supervision sources of a shared fingerprint backbone.

\section{Method}
The proposed UoU paradigm is built around one central design principle: the reusable foundation backbone, rather than any single prediction head, is the primary long-term asset. Accordingly, the present implemented branch should be viewed as an initial supervised interface for learning fingerprint-aware features, while the full UoU design elevates those features into a universal representation space suitable for large-scale pretraining and downstream transfer. In this sense, UoU should be interpreted as a modeling paradigm rather than as a commitment to one exact neural architecture.

Concretely, the method can be read at three layers. At the \emph{instantiation layer}, the present system uses a convolutional-transformer encoder-decoder and query-based structured prediction as one realized branch. At the \emph{representation layer}, the same shared backbone is designed to expose patch-level and point-level tokens together with a compact global descriptor. At the \emph{foundation-model layer}, these representations are trained and adapted with a staged recipe that combines supervised cold start, large-scale weak supervision, iterative large-scale unsupervised consolidation, and scalable deployment across model configurations. The same high-level blueprint could in principle be instantiated with alternative transformer variants, token-based foundation backbones, diffusion-enhanced perception modules, or other large-model designs that preserve reusable fingerprint representations.

\subsection{Overview and Problem Formulation}
A useful fingerprint foundation model should satisfy four properties. First, it should preserve geometric faithfulness, since fingerprint utility strongly depends on orientation, local structure, and spatial correspondences. Second, it should support semantic abstraction beyond handcrafted local templates, so that the same backbone can serve multiple downstream tasks. Third, it should remain storage-aware by allowing compact token or codebook-style representations. Fourth, it should support progressive adaptation, from large-scale generic fingerprint pretraining to specialized downstream transfer and task adaptation.

A practical consequence of this philosophy is architectural openness. The paradigm should be able to absorb different large-model choices over time, including detector-style transformers, masked autoencoding encoders, token-compression architectures, diffusion-assisted restoration modules, or hybrid multi-branch systems. The present implementation therefore serves as an anchor point, not as the architectural boundary of UoU.

\paragraph{Representation Hierarchy}
The intended UoU hierarchy can be summarized as follows:
\begin{itemize}
\item \textbf{Pixel or image space:} enhanced, restored, or normalized fingerprint images.
\item \textbf{Spatial field space:} orientation field, ridge period, mask, and related dense structural cues.
\item \textbf{Token space:} semantic units that encode recurring local patterns, modality attributes, and location-aware structures.
\item \textbf{Point space:} singular points and minutiae with interpretable geometry.
\item \textbf{Global space:} compact descriptors for matching, retrieval, and recognition.
\end{itemize}
This hierarchy is important because it lets the framework accommodate both classical biometric priors and modern large-model training recipes.

\paragraph{Notation and Task Formulation}
We formulate UoU at the level of a universal fingerprint foundation backbone rather than at the level of one specific task head. Let $I$ denote a fingerprint image sampled from a fingerprint data universe $\mathcal{D}$ that may contain different sensors, qualities, capture conditions, and supervision levels. The objective of UoU is to learn a backbone
\begin{equation}
E_{\theta}: I \mapsto \mathcal{R},
\end{equation}
where $\mathcal{R}$ is a multi-level representation hierarchy rather than a single feature tensor.

Concretely, we write
\begin{equation}
\mathcal{R} = \{\mathbf{r}^{img}, \mathbf{r}^{field}, \mathbf{r}^{tok}, \mathbf{r}^{pt}, \mathbf{r}^{glob}\},
\end{equation}
where $\mathbf{r}^{img}$ denotes image-level outputs such as enhancement or restoration targets, $\mathbf{r}^{field}$ denotes dense structural fields such as orientation or mask cues, $\mathbf{r}^{tok}$ denotes token-level semantic representations, $\mathbf{r}^{pt}$ denotes point-level structures such as core, delta, or minutiae, and $\mathbf{r}^{glob}$ denotes compact global descriptors for matching or recognition.

Different downstream tasks consume different subsets of this representation hierarchy. Let $\tau \in \mathcal{T}$ denote a downstream task such as structural parsing, matching, pose estimation, dense registration, spoof detection, or enhancement. Then each task is implemented through a task-specific adaptation map
\begin{equation}
H_{\tau}: \mathcal{R}_{\tau} \mapsto y_{\tau},
\end{equation}
where $\mathcal{R}_{\tau} \subseteq \mathcal{R}$ is the subset of representation levels used by task $\tau$, and $y_{\tau}$ is the task output.

Under this view, the central UoU hypothesis is that a single backbone $E_{\theta}$ can be pretrained and refined so that the induced representation hierarchy remains broadly useful across heterogeneous tasks and acquisition conditions. The same design may further appear as a family of model scales for different cost budgets and downstream specializations. The current repository corresponds only to one specific case of this general formulation, where $\tau$ is a structured set-prediction task and $\mathcal{R}_{\tau}$ is dominated by point-aware decoder features.

Training also follows a hierarchy. In the most general case, the universal backbone is optimized by a combination of cold-start supervision, weakly supervised expansion, and unsupervised consolidation:
\begin{equation}
\mathcal{L} = \mathcal{L}_{cold} + \lambda_{weak}\mathcal{L}_{weak} + \lambda_{unsup}\mathcal{L}_{unsup},
\end{equation}
where $\mathcal{L}_{cold}$ represents high-quality structured supervision on carefully annotated fingerprint data, $\mathcal{L}_{weak}$ represents large-scale weak or noisy supervision from broader fingerprint collections, and $\mathcal{L}_{unsup}$ denotes large-scale unsupervised consolidation objectives used during iterative representation stabilization. The exact architecture used to instantiate $E_{\theta}$ may vary; what remains fixed is the universal foundation-model objective and the representation hierarchy it is expected to support.

\subsection{Prototype Implementation}
\paragraph{Problem Setup}
Let a fingerprint image be denoted as $I \in \mathbb{R}^{H \times W}$. The present implemented branch focuses on fingerprint-aware structured prediction with point-level supervision as one concrete instantiation. Each annotated entity is represented by spatial coordinates and orientation-derived attributes. After normalization, a fingerprint is associated with an unordered target set
\begin{equation}
\mathcal{Y} = \{(c_i, \mathbf{p}_i)\}_{i=1}^{N},
\end{equation}
where $c_i$ is the structural category and $\mathbf{p}_i$ contains normalized geometry. In the present implementation, the geometry is encoded as $(x, y, \sin\theta, \cos\theta)$ after normalization to $[0,1]$.

\paragraph{Data Pipeline}
The prototype data loader performs four important operations. First, grayscale images are read and centrally normalized to a target resolution of $512 \times 512$. Second, structural annotations are loaded and organized into a unified point-level representation. Third, augmentation applies random left-right flipping, translation, rotation, and occasional masking of rectangular regions. Fourth, image intensity is inverted and scaled, while point attributes are normalized to a bounded coordinate-angle representation.

\begin{figure*}[t]
  \centering
  \includegraphics[width=0.95\textwidth]{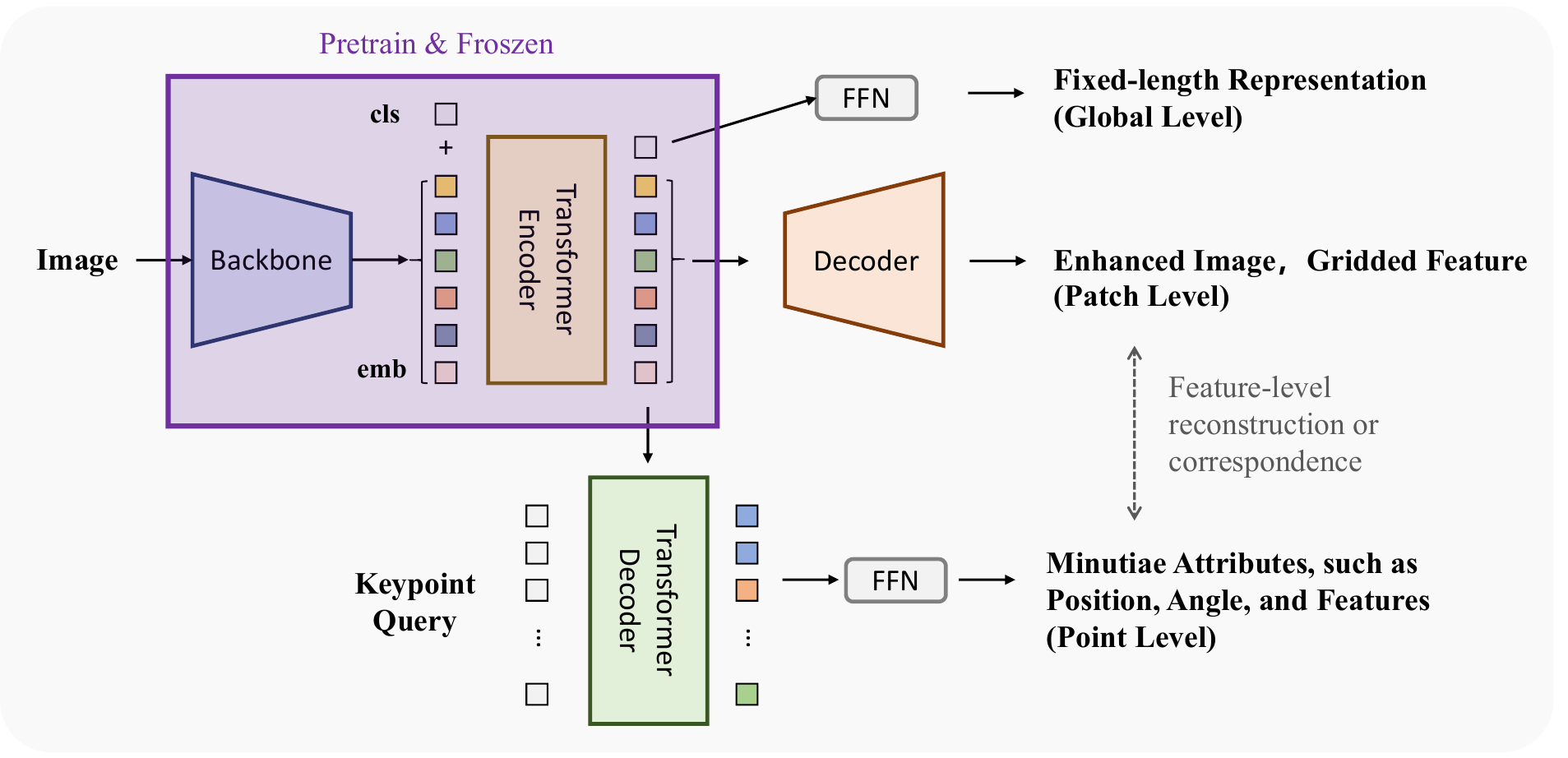}
  \caption{Conceptual overview of the UoU paradigm. The figure situates the present transformer-based structured-prediction branch within a broader fingerprint foundation-model framework. Beyond the initial branch, UoU is designed to couple tokenized representations, enhanced images, global descriptors, and reconstruction or correspondence learning within one shared modeling program.}
  \label{fig:overview}
\end{figure*}

\paragraph{Model Architecture}
The present implementation combines a ResNet backbone, a channel projection layer, sine positional embeddings, learned object queries, and a transformer encoder-decoder stack. Two prediction heads operate on decoder outputs: a classification head for structural type and a regression head for normalized point attributes. Figure~\ref{fig:overview} places this implementation in the broader UoU framework. This branch provides one initial structural instantiation, whereas the overall UoU thesis remains architecture-agnostic rather than tied to any single query-based architecture.

Formally, the baseline follows
\begin{align}
\mathbf{F} &= E(I), \\
\mathbf{Z} &= T_{\text{enc}}(\mathbf{F} + \mathbf{P}), \\
\mathbf{H} &= T_{\text{dec}}(\mathbf{Q}, \mathbf{Z}), \\
\hat{\mathcal{Y}} &= \{(\hat{c}_j, \hat{\mathbf{p}}_j)\}_{j=1}^{M},
\end{align}
where $E(\cdot)$ is the convolutional encoder, $\mathbf{P}$ is the positional encoding, $\mathbf{Q}$ denotes learned queries, and $M$ is the fixed query budget.

From the perspective of future reuse, the most important object is the latent representation between $E(\cdot)$ and the task heads. In the present implementation, this latent space is consumed by one structured-prediction branch. In the full UoU framework, the same latent space is intended to support tokenization, reconstruction, correspondence learning, and global descriptor extraction.

\paragraph{Training Objective}
The prototype adopts Hungarian matching to align predicted queries with ground-truth structural entities. This is a natural choice because the target is a variable-cardinality set rather than an ordered sequence. The implementation combines classification and coordinate regression costs for matching, and supervises class prediction together with cardinality-aware statistics during optimization. Conceptually, the objective can be summarized as
\begin{equation}
\mathcal{L} = \lambda_{\text{cls}}\mathcal{L}_{\text{cls}} + \lambda_{\text{box}}\mathcal{L}_{\text{geom}} + \lambda_{\text{card}}\mathcal{L}_{\text{card}}.
\end{equation}

\subsection{Foundation-Model Design}
\paragraph{Core Positioning}
The central thesis of UoU is that fingerprint understanding should be organized around a shared domain backbone pretrained at scale and adapted across downstream tasks. Under this view, the current structured-prediction branch is not the endpoint of the method, but one implemented head attached to a more general backbone that can later support enhancement, repair, alignment, matching, identification, anti-spoofing, and other tasks. What matters is therefore the reuse of fingerprint-aware representations rather than commitment to one particular branch design.

This framing is especially suitable for fingerprint data because the domain admits multiple semantically meaningful supervisory channels:
\begin{itemize}
\item image-level signals such as quality and modality,
\item patch-level ridge texture and local pattern tokens,
\item point-level singular points and minutiae,
\item field-level attributes such as orientation flow and ridge period,
\item global descriptors for retrieval and recognition.
\end{itemize}

\begin{figure}[t]
  \centering
  \includegraphics[width=\columnwidth]{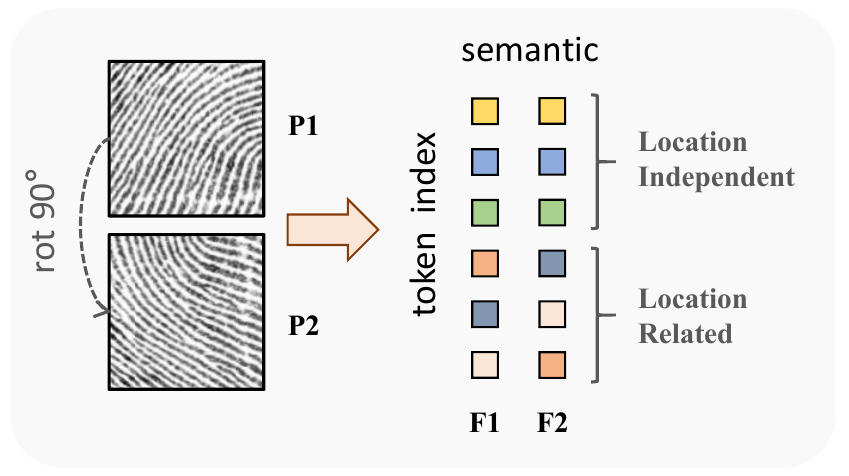}
  \caption{Tokenized representation strategy in UoU. The figure illustrates a tokenizer family that separates location-independent semantic tokens from location-related geometric tokens. This design is intended to improve representation reuse across downstream tasks while supporting more compact codebook-style storage.}
  \label{fig:token}
\end{figure}

\paragraph{Tokenized Multi-Level Representation}
UoU adopts a token-centric formulation in which fingerprint information is organized into semantically structured units rather than only dense feature maps. Figure~\ref{fig:token} summarizes this direction. Some tokens are location-independent and capture recurring local texture or modality patterns. Others are location-related and preserve geometry, such as orientation fields, periodic structures, and minutiae-linked cues.

Under this design, UoU would expose at least four representation levels:
\begin{itemize}
\item \textbf{Image level:} enhanced or restored fingerprint images.
\item \textbf{Patch level:} gridded descriptors for local ridge and texture patterns.
\item \textbf{Point level:} structured tokens linked to minutiae or singular points.
\item \textbf{Global level:} fixed-length descriptors for retrieval, matching, or identification.
\end{itemize}

This organization provides a clear path for downstream adaptation. A downstream task may consume one representation level or several jointly, while keeping the backbone shared.

\paragraph{Tokenizer Family}
One of the most distinctive aspects of UoU is the move from a monolithic feature map to a family of fingerprint-aware tokenizers. Rather than forcing all information into one homogeneous latent grid, UoU can allocate different token types to different forms of evidence:
\begin{itemize}
\item \textbf{Location-independent tokens} for recurring local texture primitives, modality traits, or latent pattern clusters.
\item \textbf{Location-related tokens} for geometry-sensitive structures such as orientation flow, ridge period, and singularity-aware regions.
\item \textbf{Point-conditioned tokens} anchored by predicted or annotated core, delta, and minutiae structures.
\item \textbf{Global summary tokens} for fixed-length identification-oriented descriptors.
\end{itemize}
This tokenizer-family view reconciles interpretability and scalability. It preserves semantically meaningful structure while still supporting large-scale unsupervised or weakly supervised scaling and compact representation.

\paragraph{Model Family and Scaling Strategy}
An important consequence of the UoU formulation is that the project should not be understood as a single frozen network instance. Instead, the shared backbone may appear in multiple model configurations with different parameter budgets, token granularities, and decoder complexity levels. A smaller UoU variant may prioritize efficient deployment for embedded recognition or mobile settings, whereas a larger variant may emphasize richer token semantics, stronger geometric reasoning, and broader multi-task transfer.

This scaling view is important for fingerprint applications because deployment conditions vary dramatically across laboratories, devices, and operational constraints. A universal fingerprint foundation model should therefore support not only task transfer, but also \emph{capacity transfer}: the same design philosophy should remain valid when instantiated as lightweight, standard, or large-capacity versions. In this sense, the reusable backbone is shared conceptually across the family even when the concrete model size changes.

\paragraph{Task Head Ecosystem}
Under the UoU paradigm, the shared backbone is not tied to a single decoder or prediction head. Instead, it is intended to support an ecosystem of heads, each aligned with a different downstream objective:
\begin{itemize}
\item structural heads for sparse geometric entity extraction,
\item dense heads for masks, orientation fields, and ridge-period estimation,
\item reconstruction heads for enhancement, repair, and image restoration,
\item representation heads for matching, retrieval, and identity discrimination,
\item auxiliary heads for anti-spoofing or quality assessment.
\end{itemize}
This perspective is important because it makes the present implemented branch a special case of a much broader multi-head system rather than the main terminal design. It also keeps the framework adaptable as stronger large-model architectures emerge.

More broadly, the task-head ecosystem is the mechanism through which UoU becomes a unified domain framework rather than a single-task recognizer. Different application scenarios may attach different heads, but the scientific claim is that they should still benefit from a shared backbone and a shared representational grammar. This is the point at which fingerprint-specific priors, large-scale unsupervised or weakly supervised scaling, and downstream specialization become part of the same research program.

\paragraph{Backbone Reuse and Downstream Adaptation}
We therefore formulate the shared backbone as the invariant core of the paradigm. In the current stage, it supports one structured-prediction branch. In later stages, the same backbone can be adapted with lightweight heads, cross-attention decoders, reconstruction modules, or contrastive objectives. This makes UoU a domain foundation model for fingerprint analysis rather than a task-isolated detector.

More specifically, downstream adaptation can proceed in at least two modes. In a lightweight mode, the backbone is frozen or partially frozen and only small task heads are optimized for target applications. In a deeper adaptation mode, the backbone is jointly fine-tuned with downstream supervision to inject task-specific discrimination while preserving the universal structure learned during earlier cold-start and iterative large-scale training stages. This flexibility is one of the main reasons to formulate the project around a reusable backbone rather than a task-isolated detector.

\paragraph{Task Head Extensions}
Within the UoU design, the following task heads fit naturally into the framework:
\begin{itemize}
\item a mask or segmentation head for foreground and structure-aware parsing,
\item a feature head for orientation field, ridge period, or gridded local descriptors,
\item an enhancement or restoration head for image reconstruction,
\item a fixed-length representation head for matching and retrieval,
\item multi-task supervision that couples reconstruction, discrimination, and spatial self-supervision.
\end{itemize}

Taken together, the tokenizer family, model scaling strategy, and task-head ecosystem define the main technical identity of UoU. The present study instantiates one structured-prediction branch, while the complete UoU design extends that branch into a reusable foundation-model framework for fingerprint intelligence.

\subsection{Training Strategy}
\begin{figure}[t]
  \centering
  \includegraphics[width=\columnwidth]{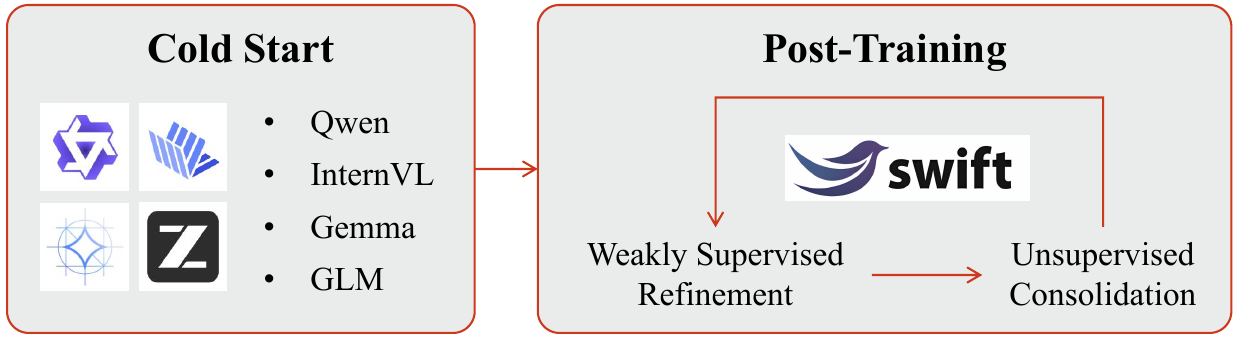}
  \caption{Training roadmap of UoU. The figure summarizes a three-stage process consisting of supervised cold start on precise annotations, large-scale weakly supervised refinement, and large-scale unsupervised consolidation, with the latter two stages intended to iterate during large-scale training. Its design is informed by representative ingredients from transformer set prediction~\cite{carion2020detr}, masked or self-supervised pretraining~\cite{he2022mae,dosovitskiy2021vit}, compact fingerprint representation learning~\cite{engelsma2019deepprint,pan2025fixedlengthdense}, and open multimodal model or adaptation ecosystems such as Qwen-VL~\cite{bai2023qwenvl}, InternVL~\cite{chen2023internvl,chen2024internvl25}, Gemma~\cite{gemma2025gemma3}, GLM-V~\cite{glmv2025glm45v}, and SWIFT / ms-swift~\cite{swiftgithub}.}
  \label{fig:train}
\end{figure}

Figure~\ref{fig:train} outlines the staged UoU training roadmap. Its motivation is aligned with a broader recent trend: large-scale open model ecosystems increasingly separate backbone initialization, data scaling, task adaptation, and efficient specialization into modular stages. Representative examples include Qwen-VL~\cite{bai2023qwenvl}, InternVL~\cite{chen2023internvl,chen2024internvl25}, Gemma~\cite{gemma2025gemma3}, GLM-V~\cite{glmv2025glm45v}, and adaptation frameworks such as SWIFT / ms-swift~\cite{swiftgithub}. UoU follows a similar systems view, but specializes it to fingerprint data, fingerprint geometry, and biometric supervision. We summarize the resulting roadmap in three phases.

\textbf{Phase 1: supervised cold start with precise annotations.}
Use a relatively small but high-quality fingerprint set with accurate labels to initialize the shared backbone and establish reliable biometric semantics. Typical supervision may include sparse structural entities, foreground masks, orientation fields, ridge-period cues, pose labels, or carefully curated matching labels. The goal of this stage is to teach the model the correct fingerprint grammar before exposing it to noisier large-scale data.

\textbf{Phase 2: large-scale weakly supervised refinement.}
Expand the backbone with much larger collections carrying weak, noisy, automatically generated, or task-dependent labels, for example field fingerprints, partial impressions, cross-sensor samples, or low-quality operational data. This stage broadens domain coverage and strengthens robustness under real acquisition variability. The present implemented branch is naturally compatible with this expansion strategy.

\textbf{Phase 3: large-scale unsupervised consolidation.}
Adapt and consolidate the shared backbone with large-scale unsupervised objectives on broader fingerprint corpora, for example through masked modeling, correspondence learning, reconstruction, consistency under transformations, or codebook-level regularization. This phase is intended to exploit richer unlabeled resources while improving deployment robustness, cross-domain transfer, and representation stability under acquisition variability.

\paragraph{Cold-Start and Weak-Supervision Objectives}
The first two stages should not rely on generic image modeling alone. A stronger domain-specific recipe can combine:
\begin{itemize}
\item precise supervised learning on carefully annotated masks, fields, sparse points, or identity cues,
\item correspondence learning across transformed fingerprint views,
\item equivariant or invariant constraints under rotation, translation, mirroring, and partial masking,
\item weak supervision from structured domain signals such as orientation, periodicity, pose, or automatically generated pseudo-labels.
\end{itemize}
The main principle is to inject fingerprint priors early so that the learned latent space remains meaningful under realistic acquisition variations even after large-scale expansion with noisier data. This is important because fingerprint signals are not generic textures. Several intermediate structures exhibit regular Euclidean symmetries that can be exploited explicitly during training. Orientation-related fields should follow rotation-covariant and translation-equivariant behavior; ridge-period or periodic-pattern representations should remain spatially consistent under translation and predictable under rotation; local feature representations should favor rotation-equivariant and translation-equivariant behavior; and sparse structures such as minutiae type, foreground mask, or singular regions can provide structural anchors even when later-stage supervision is imperfect. In this sense, the UoU recipe is better understood as a data-quality-aware geometric training program than as generic unsupervised representation learning.

\paragraph{Iterative Unsupervised Consolidation}
After cold-start supervision and weakly supervised expansion, UoU should be further consolidated with large-scale unsupervised objectives. Suitable mechanisms include masked reconstruction, cross-view correspondence constraints, consistency under geometric transformations, hybrid restoration-plus-discrimination learning, and token-level regularization over large unlabeled corpora. Phases 2 and 3 should therefore be interpreted as an iterative loop rather than as strictly separated one-pass stages: weak supervision improves semantic structure, while unsupervised consolidation stabilizes representations and provides stronger priors for the next round of large-scale refinement. UoU is thus neither a purely annotation-free pipeline nor a conventional supervised fine-tuning recipe. It is a mixed large-scale training paradigm in which a small amount of carefully annotated data provides semantic precision, weakly labeled corpora broaden coverage, and unsupervised consolidation improves representation stability on richer operational data.

\section{Evaluation Protocol}
This paper focuses on the UoU design, prototype instantiation, and validation methodology. Accordingly, this section outlines a compact evaluation protocol for future study, rather than presenting an overextended experimental section without supporting results.

\subsection{Prototype Validation}
The first validation target is the current structured-prediction branch. It should confirm that the shared backbone can learn stable geometric parsing under realistic perturbations. Recommended metrics include:
\begin{itemize}
\item detection precision and recall under spatial tolerance,
\item localization error in pixels or normalized coordinates,
\item angular error for orientation-aware points,
\item cardinality error for predicted point counts.
\end{itemize}
This stage anchors the framework in observable implementation evidence. Even though the full UoU design is broader than the present instantiated branch, the initial branch should still demonstrate that the shared backbone can learn meaningful geometry under realistic augmentation and partial observation.

\subsection{Cross-Task Evaluation}
Once the shared backbone is reused beyond the initial structured branch, evaluation should include:
\begin{itemize}
\item downstream fine-tuning accuracy on matching or identification,
\item robustness under low quality, partial overlap, and deformation,
\item storage and retrieval efficiency for fixed-length representations,
\item transfer performance across fingerprint modalities or sensors.
\end{itemize}
The central purpose of this stage is to test whether the learned representation hierarchy generalizes beyond the supervised endpoint used to initialize the present implementation.

\subsection{Comparison and Ablation Principles}
Important comparisons and ablations should study:
\begin{itemize}
\item set prediction versus heatmap or dense-regression alternatives,
\item the impact of augmentation, especially masking and rotation,
\item shared-backbone transfer to downstream tasks,
\item token granularity and codebook design,
\item cold-start-only training versus weak supervision plus unsupervised consolidation.
\end{itemize}
These analyses are necessary because the main scientific claims of UoU concern reusable structure, not merely single-task accuracy. A convincing study should therefore show which aspects of the framework contribute to transfer, robustness, compactness, and deployment flexibility.

\subsection{Minimal Evaluation Matrix}
A compact evaluation matrix for future experiments is as follows:
\begin{itemize}
\item \textbf{Prototype branch:} structured prediction under point-level supervision, measured by precision, recall, localization error, and angular error.
\item \textbf{Pretrained backbone:} patch or global tokens for matching and retrieval, measured by verification accuracy, ROC, and storage cost.
\item \textbf{Dense structural head:} field-level features for mask or orientation estimation, measured by pixel accuracy and field consistency.
\item \textbf{Downstream adapted model:} mixed-level representation for recognition or spoof detection, measured by task accuracy and robustness.
\end{itemize}
This matrix is intentionally structured around representation levels rather than around isolated scripts. It reflects the central thesis of the paper: the quality of a universal fingerprint foundation model should be judged by how consistently one shared backbone supports several distinct but related task families.

\subsection{Recommended Comparison Groups}
A mature experimental section should compare against several categories of methods rather than only against one or two direct baselines:
\begin{itemize}
\item classical fingerprint pipelines with handcrafted structure modeling,
\item task-specific deep models for structural parsing or enhancement,
\item fixed-length representation models such as DeepPrint-style systems,
\item pose or alignment-specific models,
\item dense registration or deformation-field estimators,
\item ablated UoU variants with reduced supervision or reduced representation levels.
\end{itemize}
This comparison design is important because a foundation-model paper should not be evaluated only against one architectural baseline. It should instead be compared against strong specialized alternatives in each major downstream family.

\subsection{Key Scientific Questions}
The experimental section should ultimately answer a small number of high-value questions:
\begin{itemize}
\item Does a shared fingerprint foundation backbone improve transfer across heterogeneous downstream tasks?
\item Which representation level is most useful for which class of downstream applications?
\item How much benefit comes from large-scale weak supervision and unsupervised consolidation compared with a cold-start-only model?
\item Does iterative training between Phase 2 and Phase 3 outperform a one-pass schedule, and if so, how many refinement cycles are beneficial before returns diminish?
\item What scale of data, and what mixture of fingerprint types, sensors, qualities, or capture conditions, are required for a fingerprint foundation model to become broadly useful?
\item How do performance, parameter count, and training time relate to one another in the fingerprint domain, and do clear scaling trends emerge as UoU grows in capacity?
\item Does architectural openness help, that is, can the same paradigm remain effective under more than one backbone family?
\item Can the shared backbone retain robustness under low quality, partial overlap, large pose variation, and dense distortion?
\end{itemize}

\section{Discussion}
Beyond the initial prototype, UoU naturally extends to several methodological directions that remain consistent with the same foundation-model design.

\subsection{Robust Perception Under Harsh Acquisition Conditions}
One direct extension is to strengthen representation learning for weak texture, strong noise, incomplete fingerprints, and small-overlap matching conditions. In the UoU view, this is not merely a matter of stronger denoising. It requires the shared backbone to preserve identity-sensitive structure even when local evidence is missing, corrupted, or weakly observable. This motivates more robust token formation, better multimodal fusion, and greater tolerance to local corruption.

\subsection{From 2D Alignment Priors to 3D Deformation Reasoning}
Conventional rigid or elastic 2D alignment assumptions are insufficient for real contact dynamics. A stronger extension is to move toward 3D deformation-aware modeling that explicitly reasons about compression, stretching, and bending during fingerprint acquisition. Within UoU, this extension would upgrade deformation reasoning from a downstream auxiliary capability into a first-class representation component, potentially influencing token design, correspondence learning, and downstream registration heads.

\subsection{Unified Multi-Task Fingerprint Intelligence}
Another major extension is to combine enhancement, repair, generation, pre-alignment, alignment, recognition, and anti-spoofing within one shared framework. In this setting, task diversity is not a nuisance but a training asset: multi-task supervision can force the backbone to capture shared fingerprint structure that is useful across applications. This direction also reinforces the central UoU claim that fingerprint tasks should be treated as coordinated views of one domain framework rather than as isolated products of separate pipelines.

\subsection{Data Scaling and Benchmark Ecology}
Future progress depends not only on model design but also on data and evaluation practice. A mature UoU benchmark program should include standardized sample organization, reproducible protocols, fair baselines, and clearer separation between real and synthetic training resources. For a universal fingerprint foundation model, benchmark design is part of the method itself: scaling laws, transfer behavior, and model-family comparisons are only meaningful if the data and protocol layer is designed with equal care.

\subsection{Prototype Status and Reproducibility Notes}
We briefly clarify the present scope of the prototype system and its role in the overall manuscript.

\begin{itemize}
\item The available implementation provides an initial trainable branch for fingerprint-aware structured prediction and serves as an executable entry point to the broader UoU design.
\item The full UoU manuscript covers a larger foundation-model program, including geometry-aware large-scale training, tokenized multi-level representations, multi-task heads, and model-family scaling beyond the present instantiated branch.
\item A portion of the baseline implementation is publicly available in the project repository, which supports early reproducibility and future extension of the framework.
\item A submission-ready empirical version should further consolidate data interfaces, training configuration, and downstream task heads so that the broader UoU program can be evaluated under one unified protocol.
\end{itemize}

\subsection{Broader Implications}
The main scientific opportunity of UoU lies in shifting fingerprint research from isolated algorithmic modules toward a unified representation paradigm. If successful, this direction could reduce duplicated engineering across tasks and make domain knowledge reusable at scale. The shared backbone would become the carrier of ridge texture, singular structure, geometric relations, and global identity cues, while the overall framework would support diverse downstream objectives and deployment budgets.

Equally important, the proposed framing gives the work a coherent narrative arc. The present branch is not merely a narrow task model; it is the initial supervised interface through which a broader fingerprint foundation model is instantiated and examined. This perspective helps align the available implementation, future training objectives, and eventual application value into one coherent research program.

From a manuscript-design standpoint, this also clarifies the role of the implemented branch in the overall paper. The present implementation demonstrates one instantiated branch of the UoU design, while the manuscript defines the larger foundation-model system into which that branch is integrated.

A second important implication is methodological openness. If the core research claim were tied to a particular detector or backbone, the paper would risk becoming obsolete as architectures evolve. By instead centering the contribution on a universal fingerprint foundation-model paradigm and its representation hierarchy, the manuscript preserves long-term relevance: stronger transformers, tokenizers, diffusion modules, or hybrid architectures can all be viewed as future instantiations rather than contradictions of the framework.

More broadly, the UoU perspective suggests that fingerprint research may benefit from the same transition already observed in general vision and language: from pipeline-centric optimization toward reusable large-scale training, scalable adaptation, and family-level modeling. The difference is that fingerprint intelligence is unusually structured, geometry-sensitive, and supervision-rich. That combination makes the domain especially suitable for a domain-specific foundation-model framework, provided that future work converts the present design into rigorous empirical evidence.

\subsection{Limitations}
This work has several current limitations.

\begin{itemize}
\item The manuscript does not report quantitative claims beyond what is supported by the present implementation.
\item The overall UoU design is broader than the present codebase; in particular, the tokenizer family, multi-task heads, model-scaling strategy, and full three-stage training pipeline are not yet fully realized.
\item Bibliographic coverage is representative rather than exhaustive, and a submission-ready version should further expand the fingerprint-specific comparison set.
\item The evaluation protocol is presently defined at the methodology level and should be updated with finalized datasets, metrics, reproducible training scripts, and downstream task heads.
\item Several scientific claims of UoU remain to be validated through cross-task experiments, model-scale comparisons, and transfer studies on heterogeneous fingerprint data.
\end{itemize}

\section{Conclusion}
This paper formulates UoU as a universal fingerprint foundation-model paradigm built around reusable domain representations rather than isolated task solvers. The central claim is that fingerprint intelligence should be organized through a shared representation hierarchy spanning structural fields, semantic tokens, point-level entities, and compact global descriptors, so that enhancement, alignment, matching, registration, and related tasks can be treated as downstream realizations of one common backbone.

Within this view, large-scale unsupervised learning is the dominant scaling substrate of the framework, but it is not the whole training story by itself. UoU instead follows a mixed large-scale recipe: a supervised cold start establishes reliable fingerprint semantics, large-scale weak supervision broadens domain coverage, and large-scale unsupervised consolidation stabilizes representation geometry and transfer under heterogeneous acquisition conditions. Crucially, the latter two stages are most naturally interpreted as an iterative loop rather than as a rigid one-pass sequence.

The present study provides an initial implemented branch that makes this framework concrete while leaving room for broader model families, larger-scale data mixtures, and richer downstream heads. UoU should therefore be understood not as a single encoder instance, but as a domain-specific foundation-model program for fingerprint analysis; the acronym is intentionally tied to the phrase ``\textbf{U}niversal fingerprint foundation model based \textbf{o}n large-scale \textbf{U}nsupervised learning.'' We hope this formulation helps shift fingerprint research from task-isolated optimization toward scalable, reusable, and empirically grounded foundation modeling.

\bibliographystyle{ieeetr}
\bibliography{refs}

@inproceedings{carion2020detr,
author       = {Nicolas Carion and
                  Francisco Massa and
                  Gabriel Synnaeve and
                  Nicolas Usunier and
                  Alexander Kirillov and
                  Sergey Zagoruyko},
  editor       = {Andrea Vedaldi and
                  Horst Bischof and
                  Thomas Brox and
                  Jan{-}Michael Frahm},
  title        = {End-to-End Object Detection with Transformers},
  booktitle    = {Computer Vision - {ECCV} 2020 - 16th European Conference, Glasgow,
                  UK, August 23-28, 2020, Proceedings, Part {I}},
  series       = {Lecture Notes in Computer Science},
  volume       = {12346},
  pages        = {213--229},
  publisher    = {Springer},
  year         = {2020},
  url          = {https://doi.org/10.1007/978-3-030-58452-8\_13},
  doi          = {10.1007/978-3-030-58452-8\_13},
}

@inproceedings{vaswani2017attention,
  author       = {Ashish Vaswani and
                  Noam Shazeer and
                  Niki Parmar and
                  Jakob Uszkoreit and
                  Llion Jones and
                  Aidan N. Gomez and
                  Lukasz Kaiser and
                  Illia Polosukhin},
  editor       = {Isabelle Guyon and
                  Ulrike von Luxburg and
                  Samy Bengio and
                  Hanna M. Wallach and
                  Rob Fergus and
                  S. V. N. Vishwanathan and
                  Roman Garnett},
  title        = {Attention is All you Need},
  booktitle    = {Advances in Neural Information Processing Systems 30: Annual Conference
                  on Neural Information Processing Systems 2017, December 4-9, 2017,
                  Long Beach, CA, {USA}},
  pages        = {5998--6008},
  year         = {2017},
url          = {https://proceedings.neurips.cc/paper/2017/hash/3f5ee243547dee91fbd053c1c4a845aa-Abstract.html},
}

@inproceedings{he2016resnet,
  author       = {Kaiming He and
                  Xiangyu Zhang and
                  Shaoqing Ren and
                  Jian Sun},
  title        = {Deep Residual Learning for Image Recognition},
  booktitle    = {2016 {IEEE} Conference on Computer Vision and Pattern Recognition,
                  {CVPR} 2016, Las Vegas, NV, USA, June 27-30, 2016},
  pages        = {770--778},
  publisher    = {{IEEE} Computer Society},
  year         = {2016},
url          = {https://doi.org/10.1109/CVPR.2016.90},
  doi          = {10.1109/CVPR.2016.90},
}

@inproceedings{dosovitskiy2021vit,
  author       = {Alexey Dosovitskiy and
                  Lucas Beyer and
                  Alexander Kolesnikov and
                  Dirk Weissenborn and
                  Xiaohua Zhai and
                  Thomas Unterthiner and
                  Mostafa Dehghani and
                  Matthias Minderer and
                  Georg Heigold and
                  Sylvain Gelly and
                  Jakob Uszkoreit and
                  Neil Houlsby},
  title        = {An Image is Worth 16x16 Words: Transformers for Image Recognition
                  at Scale},
  booktitle    = {9th International Conference on Learning Representations, {ICLR} 2021,
                  Virtual Event, Austria, May 3-7, 2021},
  publisher    = {OpenReview.net},
  year         = {2021},
  url          = {https://openreview.net/forum?id=YicbFdNTTy},
}

@inproceedings{he2022mae,
author       = {Kaiming He and
                  Xinlei Chen and
                  Saining Xie and
                  Yanghao Li and
                  Piotr Doll{\'{a}}r and
                  Ross B. Girshick},
  title        = {Masked Autoencoders Are Scalable Vision Learners},
  booktitle    = {{IEEE/CVF} Conference on Computer Vision and Pattern Recognition,
                  {CVPR} 2022, New Orleans, LA, USA, June 18-24, 2022},
  pages        = {15979--15988},
  publisher    = {{IEEE}},
  year         = {2022},
  url          = {https://doi.org/10.1109/CVPR52688.2022.01553},
  doi          = {10.1109/CVPR52688.2022.01553},
}

@book{maltoni2009handbook,
 author       = {Davide Maltoni and
                  Dario Maio and
                  Anil K. Jain and
                  Jianjiang Feng},
  title        = {Handbook of Fingerprint Recognition, Third Edition},
  publisher    = {Springer},
  year         = {2022},
  url          = {https://doi.org/10.1007/978-3-030-83624-5},
  doi          = {10.1007/978-3-030-83624-5},
  isbn         = {978-3-030-83623-8},
}

@article{cappelli2006fvc2004,
author       = {Dario Maio and
                  Davide Maltoni and
                  Raffaele Cappelli and
                  James L. Wayman and
                  Anil K. Jain},
  editor       = {David Zhang and
                  Anil K. Jain},
  title        = {{FVC2004:} Third Fingerprint Verification Competition},
  booktitle    = {Biometric Authentication, First International Conference, {ICBA} 2004,
                  Hong Kong, China, July 15-17, 2004, Proceedings},
  series       = {Lecture Notes in Computer Science},
  volume       = {3072},
  pages        = {1--7},
  publisher    = {Springer},
  year         = {2004},
  url          = {https://doi.org/10.1007/978-3-540-25948-0\_1},
  doi          = {10.1007/978-3-540-25948-0\_1},
}

@article{tang2017fingernet,
author       = {Yao Tang and
                  Fei Gao and
                  Jufu Feng and
                  Yuhang Liu},
  title        = {FingerNet: An unified deep network for fingerprint minutiae extraction},
  booktitle    = {2017 {IEEE} International Joint Conference on Biometrics, {IJCB} 2017,
                  Denver, CO, USA, October 1-4, 2017},
  pages        = {108--116},
  publisher    = {{IEEE}},
  year         = {2017},
  url          = {https://doi.org/10.1109/BTAS.2017.8272688},
  doi          = {10.1109/BTAS.2017.8272688},
}

@article{nguyen2018minutiaenet,
author       = {Dinh{-}Luan Nguyen and
                  Kai Cao and
                  Anil K. Jain},
  title        = {Robust Minutiae Extractor: Integrating Deep Networks and Fingerprint
                  Domain Knowledge},
  booktitle    = {2018 International Conference on Biometrics, {ICB} 2018, Gold Coast,
                  Australia, February 20-23, 2018},
  pages        = {9--16},
  publisher    = {{IEEE}},
  year         = {2018},
  url          = {https://doi.org/10.1109/ICB2018.2018.00013},
  doi          = {10.1109/ICB2018.2018.00013},
}

@article{engelsma2019deepprint,
author       = {Joshua J. Engelsma and
                  Kai Cao and
                  Anil K. Jain},
  title        = {Learning a Fixed-Length Fingerprint Representation},
  journal      = {{IEEE} Trans. Pattern Anal. Mach. Intell.},
  volume       = {43},
  number       = {6},
  pages        = {1981--1997},
  year         = {2021},
  url          = {https://doi.org/10.1109/TPAMI.2019.2961349},
  doi          = {10.1109/TPAMI.2019.2961349},
}

@article{tang2017latentfcn,
author       = {Yao Tang and
                  Fei Gao and
                  Jufu Feng},
  title        = {Latent fingerprint minutia extraction using fully convolutional network},
  booktitle    = {2017 {IEEE} International Joint Conference on Biometrics, {IJCB} 2017,
                  Denver, CO, USA, October 1-4, 2017},
  pages        = {117--123},
  publisher    = {{IEEE}},
  year         = {2017},
  url          = {https://doi.org/10.1109/BTAS.2017.8272689},
  doi          = {10.1109/BTAS.2017.8272689},
}

@article{pan2024dmd,
author       = {Zhiyu Pan and
                  Yongjie Duan and
                  Xiongjun Guan and
                  Jianjiang Feng and
                  Jie Zhou},
  title        = {Latent Fingerprint Matching via Dense Minutia Descriptor},
  booktitle    = {{IEEE} International Joint Conference on Biometrics, {IJCB} 2024,
                  Buffalo, NY, USA, September 15-18, 2024},
  pages        = {1--10},
  publisher    = {{IEEE}},
  year         = {2024},
  url          = {https://doi.org/10.1109/IJCB62174.2024.10744445},
  doi          = {10.1109/IJCB62174.2024.10744445},
}

@article{pan2025fixedlengthdense,
author       = {Zhiyu Pan and
                  Xiongjun Guan and
                  Yongjie Duan and
                  Jianjiang Feng and
                  Jie Zhou},
  title        = {Fixed-Length Dense Fingerprint Representation With Alignment and Robust
                  Enhancement},
  journal      = {{IEEE} Trans. Inf. Forensics Secur.},
  volume       = {21},
  pages        = {1751--1765},
  year         = {2026},
  url          = {https://doi.org/10.1109/TIFS.2026.3658990},
  doi          = {10.1109/TIFS.2026.3658990},
}

@article{guan2025underpose,
 author       = {Xiongjun Guan and
                  Zhiyu Pan and
                  Jianjiang Feng and
                  Jie Zhou},
  title        = {Finger Pose Estimation for Under-Screen Fingerprint Sensor},
  journal      = {{IEEE} Trans. Inf. Forensics Secur.},
  volume       = {20},
  pages        = {12739--12753},
  year         = {2025},
  url          = {https://doi.org/10.1109/TIFS.2025.3636671},
  doi          = {10.1109/TIFS.2025.3636671},
}

@article{guan2024jointpose,
author       = {Xiongjun Guan and
                  Zhiyu Pan and
                  Jianjiang Feng and
                  Jie Zhou},
  title        = {Joint Identity Verification and Pose Alignment for Partial Fingerprints},
  journal      = {{IEEE} Trans. Inf. Forensics Secur.},
  volume       = {20},
  pages        = {249--263},
  year         = {2025},
  url          = {https://doi.org/10.1109/TIFS.2024.3516566},
  doi          = {10.1109/TIFS.2024.3516566},
}

@article{guan2024distortionfield,
author       = {Xiongjun Guan and
                  Yongjie Duan and
                  Jianjiang Feng and
                  Jie Zhou},
  title        = {Regression of Dense Distortion Field From a Single Fingerprint Image},
  journal      = {{IEEE} Trans. Inf. Forensics Secur.},
  volume       = {18},
  pages        = {4377--4390},
  year         = {2023},
  url          = {https://doi.org/10.1109/TIFS.2023.3296310},
  doi          = {10.1109/TIFS.2023.3296310},
}

@article{guan2024pdrnet,
author       = {Xiongjun Guan and
                  Jianjiang Feng and
                  Jie Zhou},
  title        = {Phase-Aggregated Dual-Branch Network for Efficient Fingerprint Dense
                  Registration},
  journal      = {{IEEE} Trans. Inf. Forensics Secur.},
  volume       = {19},
  pages        = {5712--5724},
  year         = {2024},
  url          = {https://doi.org/10.1109/TIFS.2024.3403507},
  doi          = {10.1109/TIFS.2024.3403507},
}

@article{engelsma2022printsgan,
author       = {Joshua J. Engelsma and
                  Steven A. Grosz and
                  Anil K. Jain},
  title        = {PrintsGAN: Synthetic Fingerprint Generator},
  journal      = {{IEEE} Trans. Pattern Anal. Mach. Intell.},
  volume       = {45},
  number       = {5},
  pages        = {6111--6124},
  year         = {2023},
  url          = {https://doi.org/10.1109/TPAMI.2022.3204591},
  doi          = {10.1109/TPAMI.2022.3204591},
}

@article{bai2023qwenvl,
title={Qwen-VL: A Versatile Vision-Language Model for Understanding, Localization, Text Reading, and Beyond}, 
      author={Jinze Bai and Shuai Bai and Shusheng Yang and Shijie Wang and Sinan Tan and Peng Wang and Junyang Lin and Chang Zhou and Jingren Zhou},
      year={2023},
      eprint={2308.12966},
      archivePrefix={arXiv},
      primaryClass={cs.CV},
      url={https://arxiv.org/abs/2308.12966}, 
}

@article{chen2023internvl,
 title={InternVL: Scaling up Vision Foundation Models and Aligning for Generic Visual-Linguistic Tasks}, 
      author={Zhe Chen and Jiannan Wu and Wenhai Wang and Weijie Su and Guo Chen and Sen Xing and Muyan Zhong and Qinglong Zhang and Xizhou Zhu and Lewei Lu and Bin Li and Ping Luo and Tong Lu and Yu Qiao and Jifeng Dai},
      year={2024},
      eprint={2312.14238},
      archivePrefix={arXiv},
      primaryClass={cs.CV},
      url={https://arxiv.org/abs/2312.14238}, 
}

@article{chen2024internvl25,
 title={InternVideo2.5: Empowering Video MLLMs with Long and Rich Context Modeling}, 
      author={Yi Wang and Xinhao Li and Ziang Yan and Yinan He and Jiashuo Yu and Xiangyu Zeng and Chenting Wang and Changlian Ma and Haian Huang and Jianfei Gao and Min Dou and Kai Chen and Wenhai Wang and Yu Qiao and Yali Wang and Limin Wang},
      year={2025},
      eprint={2501.12386},
      archivePrefix={arXiv},
      primaryClass={cs.CV},
      url={https://arxiv.org/abs/2501.12386}, 
}

@article{gemma2025gemma3,
  title={Gemma 3 Technical Report},
  author={Gemma Team and others},
  year={2025},
      eprint={2503.19786},
      archivePrefix={arXiv},
      primaryClass={cs.CL},
      url={https://arxiv.org/abs/2503.19786},
}

@article{glmv2025glm45v,
  title={GLM-4.5V and GLM-4.1V-Thinking: Towards Versatile Multimodal Reasoning with Scalable Reinforcement Learning}, 
  author={GLM Team and others},
  year={2026},
      eprint={2507.01006},
      archivePrefix={arXiv},
      primaryClass={cs.CV},
      url={https://arxiv.org/abs/2507.01006}, 
}

@article{swiftgithub,
  title={SWIFT:A Scalable lightWeight Infrastructure for Fine-Tuning}, 
      author={Yuze Zhao and Jintao Huang and Jinghan Hu and Xingjun Wang and Yunlin Mao and Daoze Zhang and Hong Zhang and Zeyinzi Jiang and Zhikai Wu and Baole Ai and Ang Wang and Wenmeng Zhou and Yingda Chen},
      year={2025},
      eprint={2408.05517},
      archivePrefix={arXiv},
      primaryClass={cs.CL},
      url={https://arxiv.org/abs/2408.05517}, 
}

\end{document}